\documentclass[letterpaper, 10 pt, conference]{ieeeconf}

\usepackage{mathtools}

\usepackage[T1]{fontenc}
\usepackage[latin9]{inputenc}
\usepackage{mathtools}
\usepackage{amsmath}
\usepackage{amsthm}
\usepackage{amssymb}
\usepackage{enumerate}
\usepackage{graphicx}
\usepackage[pagebackref]{hyperref}
\usepackage{array}
\usepackage{balance}
\usepackage{subfigure}
\usepackage{color}
\usepackage{lineno}
\usepackage{dirtytalk}
\usepackage{verbatim}
\usepackage[]{units}
\usepackage{booktabs}
\usepackage{xspace}
\usepackage[dvipsnames]{xcolor}
\usepackage{flushend}

\usepackage{algorithm}
\usepackage[noend]{algpseudocode}
\usepackage[section]{placeins}
\usepackage{hyperref}

\algrenewcommand\algorithmicindent{1.0em}

\usepackage{mathtools, cuted}

\makeatletter
\newtheorem*{theorem*}{Theorem}

\newtheorem{lem}{Lemma}
\newtheorem*{lem*}{Lemma}

\newcommand{\Xfree}{X_{\rm free}}

\newcommand{\Gammasol}{\Gamma_{\rm sol}}
\newcommand{\Gammafree}{\Gamma_{\rm free}}

\newcommand{\PRM}{\ensuremath{\mathrm{PRM}}\xspace}
\newcommand{\RRT}{\ensuremath{\mathrm{RRT}}\xspace}
\newcommand{\EST}{\ensuremath{\mathrm{EST}}\xspace}
\newcommand{\Astar}{${\rm A^*}$\xspace}

\newcommand{\Pclear}{\mathbb{P}(\texttt{clear})}

\IEEEoverridecommandlockouts
\begin{document}
\title{Landmark Guided Probabilistic Roadmap Queries}

\author{Brian Paden$^{\rm a,b}$, Yannik Nager$^{\rm a,b}$, and Emilio Frazzoli$^{\rm a}$
\thanks{$^a$ The authors are with The Institute for Dynamic Systems and Control at ETH (email: \{padenb,ynager,emilio.frazzoli\}@ethz.ch).}
\thanks{$^b$ The first two authors made equal contributions to this work.}}
\maketitle
\begin{abstract}
A landmark based heuristic is investigated for reducing query phase run-time of the probabilistic roadmap (\PRM) motion planning method. 
The heuristic is generated by storing minimum spanning trees from a small number of vertices within the \PRM graph and using these trees to approximate the cost of a shortest path between any two vertices of the graph. 
The intermediate step of preprocessing the graph increases the time and memory requirements of the classical motion planning technique in exchange for speeding up individual queries making the method advantageous in multi-query applications. 
This paper investigates these trade-offs on \PRM graphs constructed in randomized environments as well as a practical manipulator simulation.
We conclude that the method is preferable to Dijkstra's algorithm or the ${\rm A}^*$ algorithm with conventional heuristics in multi-query applications. 
\end{abstract}

\section{Introduction}
The probabilistic roadmap (\PRM)~\cite{kavraki1996probabilistic} is a cornerstone of robot motion planning. It is widely used in practice or as the foundation for more complex planning algorithms.
The method is divided into two phases: the \PRM graph is first constructed followed by, potentially multiple, shortest path queries on this graph to solve motion planning problems.   
For a single motion planning query, a feasibility checking subroutine executed repeatedly during \PRM construction dominates run-time.
However, once the \PRM is constructed it can be reused for multiple motion planning queries or modified slightly according to minor changes in the environment.
Applicability to multi-query problems is one of the advantages of the \PRM over tree-based planners such as Rapidly exploring Random Trees (\RRT)~\cite{lavalle2001randomized} and Expansive Space Trees (\EST)~\cite{hsu1997path} which are tailored to single-query problems.
Recent efforts have focused on fine tuning various aspects of PRM-based motion planning for real-time applications.
Highly parallelized feasibility checking using FPGAs was recently developed in~\cite{murray2016robot} to alleviate this computational bottleneck during the construction phase.
The sparse roadmap spanner was introduced in~\cite{marble2013asymptotically} to reduce memory required to store the \PRM and speed up the query phase by keeping only a sparse subgraph with near-optimality properties. 
In this paper we examine the effectiveness of a landmark based admissible heuristic for reducing the running time of the query phase of the \PRM. 
The landmark heuristic was originally developed for vehicle routing problems in road networks~\cite{goldberg2005computing} where many shortest path queries are solved on a single graph.
In theory, any amount of time spent preprocessing the graph is negligible in comparison to the time spent solving  shortest path queries if sufficiently many queries must be solved.
This observation suggests solving the all-pairs shortest path problem in order to answer each routing problem in constant time with respect to graph size.
However, the memory required to store a solution to the all-pairs shortest is prohibitive for large road networks.
The landmark heuristic provides a trade-off between memory requirements and query times by solving a small number of single-source shortest path problems and using their solutions to construct an effective heuristic for a particular graph.

This investigation is inspired by the similarities between road networks and the \PRM; multiple path queries are solved on both graphs and both graphs are, in practice, too large to store an all pairs shortest path solution in memory.
A useful feature of the landmark heuristic is that it can be used together with the sparse roadmap spanner and FPGA-based collision checking for a compounded speedup over a standard \PRM implementation.
Based on the results presented in this paper, we conclude that the landmark heuristic is effective on \PRM graphs; solving shortest path queries as much as 20 times faster than Dijkstra's algorithm and twice as fast as the Euclidean distance-based heuristic in cluttered environments.
The downside to the approach is that constructing the heuristic requires preprocessing the graph which adds to the computation time required before the PRM can be used for motion planning queries.

An overview of the motion planning problem is presented in Section \ref{sec:mp_background}, followed by a review of the build and query phases of the \PRM method.
%
%
%
Section \ref{sec:LH} introduces the landmark heuristic, discusses its admissibility and the complexity of its construction, and illustrates its utility with a simple shortest path problem.
However, to better understand the effectiveness of the landmark heuristic in general, we construct randomized environments with a quantifiable degree of clutter and run numerous motion planning queries on these environments to obtain the average case performance.
The environment construction and experimental results are presented in Section \ref{sec:experiments}.
In Section \ref{sec:jaco} we evaluate the landmark heuristic on a simulation of the Kinova Jaco robotic manipulator and find the landmark heuristic to be effective on realistic robot models.
Lastly, we conclude with a discussion of our experimental observations in Section \ref{sec:conclusion}.

\section{Motion Planning Problem}\label{sec:mp_background}
The following optimal motion planning problem will be addressed: 
Let $\Xfree$ be an open, bounded subset of $\mathbb{R}^d$, and $\Gamma$ the set of continuous \emph{curves} from $[0,1]$ to $\mathbb{R}^n$. 
Then let $\Gamma_{\rm free}$ be the subset of $\Gamma$ whose image is contained in $\Xfree$. 
The cost objective is a function $c:\Gamma \rightarrow [0,\infty)$ that assigns a cost to each curve in $\mathbb{R}^d$.
The cost function must be additive in the sense that if two curves $\gamma_{1,2} \in \Gamma$ satisfy $\gamma_1([0,1]) \subset \gamma_2([0,1])$ then $c(\gamma_1)\leq c(\gamma_2)$.  

An individual motion planning query on $\Xfree$ consists of finding a curve $\gamma^*\in \Gamma_{\rm free}$ from an initial state $x_0 \in \Xfree$ to a goal state $x_g\in \Xfree$. 
That is, $\gamma^*(0)=x_0$ and $\gamma^*(1)=x_g$. 
The subset of curves in $\Gammafree$ which satisfy these additional endpoint constraints are denoted $\Gammasol$. 
In addition to finding a curve in $\Gammasol$, we would like a curve $\gamma^*$ which approximately minimizes the cost objective,
\begin{equation}
c(\gamma^*)<\underset{\gamma\in\Gammasol}{\inf}(c(\gamma)) + \varepsilon,
\end{equation} 
for a fixed $\varepsilon>0$.
An approximate minimization is often used for two reasons: the first is that the problem may not admit a minimum, and second, without further assumptions on the cost objective and geometry of $\Xfree$ there are no practical techniques available for obtaining exact solutions when they exist.
\subsection{Probabilistic Roadmaps}
The set $\Gamma_{\rm free}$ has infinite dimension so the conventional approach to obtaining approximate solutions to motion planning problems is to construct a graph on $X_{\rm free}$ whose vertices are points in $\Xfree$. 
To avoid confusion with curves on $\Xfree$, a \emph{path} is a sequence of vertices in a graph $\{x_i\}$ such that $(x_i,x_{i+1})$ is an edge in the graph.
Curves in $\Gammafree$ are approximated using paths in the graph by associating each edge of the graph with the line segment between the two vertices making up that edge.
The \PRM method falls into this category of approximations to $\Gamma_{\rm free}$.

%
%
%
%

%
The ${\rm PRM^*}$ method~\cite{karaman2011sampling} is a popular variation of the \PRM because it generates a sparse graph with the following property: 
if $x_0$ and $x_g$ belong to a connected subset of $X_{\rm free}$, then for any fixed $\varepsilon>0$, the probability that the ${\rm PRM}^*$ graph contains a curve $\hat{\gamma} \in \Gammafree$ satisfying 
\begin{equation}
\begin{array}{c}
c(\hat{\gamma})<\underset{\gamma\in\Gamma_{\rm sol}}{\inf}(c(\gamma)) + \varepsilon, \\
\Vert \hat{\gamma}(0)-x_0 \Vert < \varepsilon,\\
\Vert \hat{\gamma}(1)-x_g\Vert < \varepsilon,
\end{array}\label{eq:optimality}
\end{equation} 
converges to $1$ as the number of vertices is increased. 

\subsection{Graph Construction Phase}
The construction phase of the $\PRM^*$ method is summarized in Algorithm \ref{alg:prm}.
The $\mathtt{nearest}(r,x,V_{\PRM})$ subroutine returns the points  $v\in V_{\PRM}\setminus\{x\}$ such that  $\Vert x- v\Vert<r$. 
The subroutine $\mathtt{ sample}(\Xfree)$ in Algorithm \ref{alg:prm} returns a randomly sampled point from the uniform distribution supported on $\Xfree$. 
The subroutine $\mathtt{ collisionFree}(x,v)$ returns true if the line segment connecting $x$ to $v$ is an element of $\Gamma_{\rm free}$ and false otherwise.  
In reference to line $2$ of Algorithm \ref{alg:prm}, $\mu$ is the Legesgue measure on $\mathbb{R}^d$, and $B_1(0)$ is the ball of radius $1$ centered at $0$.  

\begin{algorithm} 
	\begin{algorithmic}[1]
		\State $V_{\rm PRM}\gets \emptyset;\,\,E_{\rm PRM}\gets \emptyset$
		\State $r=\left((2+2/d)\left(\frac{\mu(\Xfree)}{\mu(B_1(0))}\right) \left(\frac{\log(n)}{n}\right)\right)^{1/d} $
		\For {$i=1,\dots, n$}
		\State $V_{\rm PRM} \gets V_{\rm PRM} \cup \{\mathtt{sample}(X_{\rm free})\}$ 
		\EndFor
		\For {$x\in V_{\rm PRM}$}    
		\State $U \gets \mathtt{nearest}(r,x,V_{\rm PRM})$ 
		\For{$v \in U\setminus\{x\}$} 	
		\If{$\mathtt{collisionFree}(x,v)$}         	  
		\State $E_{\rm PRM}=E_{\rm PRM}\cup\{(x,v)\}$
		\EndIf        	  
		\EndFor       
		\EndFor  
		\State \Return $(V_{\rm PRM},E_{\rm PRM})$ 
	\end{algorithmic} \caption{\label{alg:prm} ${\rm PRM^*}$ } 
\end{algorithm}

\subsection{Motion Planning Query Phase}
After construction, paths in the graph $(V_{\rm PRM},E_{\rm PRM})$ can be used to solve motion planning queries.
One subtlety is that the probability of $x_0$ and $x_g$ being present in the \PRM graph is zero. 
There are a number of practical ways to resolve this issue, but to keep the exposition as concise as possible we will simply select the nearest vertex $\tilde{x}_{0}\in V_{\rm PRM}$ to $x_0$ and $\tilde{x}_{g}\in V_{\rm PRM}$ to $x_g$ as an approximation in light of (\ref{eq:optimality}). 
%

%
Once initial and final states $\tilde{x}_{0}$ and $\tilde{x}_{g}$ are selected, the motion planning query reduces to a shortest path problem on the $\rm PRM$ graph with edge weights determined by the cost of the line segments between vertices of the graph.
%


Algorithm \ref{alg:ucs} summarizes the ${\rm A}^*$ algorithm for finding a shortest path in the \PRM graph from $\tilde{x}_0$ to $\tilde{x}_g$. 
The function ${\rm parent}:V_{\rm PRM}\rightarrow V_{\rm PRM}\cup \{\mathtt{ NULL}\}$ is used to keep track of the shortest path from $\tilde{x}_0$ to each vertex examined by the algorithm.
Initially, $\texttt{parent}$ maps all vertices of the graph to $\mathtt{ NULL}$, but is redefined in each iteration of the algorithm as shorter paths from $\tilde{x}_0$ to vertices in the graph are found.
The function $\mathtt{ label}:V_{\rm PRM}\rightarrow [0,\infty]$ maps each vertex to the cost of the shortest known path reaching that vertex from $\tilde{x}_0$. 
%
%
The function $\mathtt{label}$ initially maps all vertices to $\infty$, but is updated at each iteration with the cost of newly discovered paths.
A set $Q$ of vertices represents a priority queue. 
The distinguishing feature of the ${\rm A}^*$ algorithm is the ordering of vertices in the priority queue according to the labeled cost of the vertex plus a heuristic estimate of the remaining cost to reach the goal $h:V_{\rm PRM}\rightarrow [0,\infty] $.  
The subroutine $\mathtt{ pop}(Q)$ returns a vertex $x\in Q$ such that 
\begin{equation}\label{eq:order}
	x\in \underset{\nu \in Q}{{\rm argmin}} \left\{ \mathtt{ label}(\nu) + h(\nu) \right\}
\end{equation}  
The heuristic $h$ is called \emph{admissible} if it never overestimates the cost to reach the goal from a particular vertex. 
The ${\rm A}^*$ algorithm is guaranteed to return the shortest path from $\tilde{x}_0$ if the heuristic in equation (\ref{eq:order}) is admissible.
The $\texttt{pathToRoot}$ subroutine returns the sequence of vertices  $\{v_i\}_{i=1,...,N}$, terminating at $v_N=\tilde{x}_0$, generated by the recursion
\begin{equation}
v_{i+1}=\texttt{parent}(v_i),\quad v_1=\tilde{x}_g,
\end{equation}
If $\texttt{pathToRoot}$ is evaluated in Algorithm \ref{alg:ucs}, then its output is a shortest path from $\tilde{x}_0$ to $\tilde{x}_g$.
For graphs with nonnegative edge-weights the heuristic $h(x)=0$ for all $x\in V_{\rm PRM}$ is clearly admissible. 
In this special case, the ${\rm A}^*$ algorithm is equivalent to Dijkstra's algorithm.
However, the more closely $h$ underestimates the optimal cost from each vertex to $\tilde{x}_g$ the fewer iterations required by the ${\rm A}^*$ algorithm to find the shortest path from $\tilde{x}_0$ to $\tilde{x}_g$. 
Therefore, it is desirable to use a heuristic which estimates the optimal cost to reach the goal as closely as possible. 
%
%
\begin{algorithm} 
	\begin{algorithmic}[1]
		\State $Q\gets \tilde{x}_0;$
		\State $\mathtt{label}(\tilde{x}_0)\gets 0$
		\While {$Q\neq \emptyset$}
		\State $v\gets {\rm pop}(Q)$
		\If{$v=\tilde{x}_g$}
		\State \Return $\mathtt{pathToRoot}(x_g)$
		\EndIf
		\State $S\gets {\rm neighbors}(v)$
		\For $w\in S$
		\If{$\mathtt{label}(v)+\mathtt{cost}(v,w)<\mathtt{label}(w)$}
		\State $\mathtt{label}(w)\gets \mathtt{label}(v)+\mathtt{cost}(v,w)$
		\State $\mathtt{parent}(w)\gets v$
		\State $Q\gets Q\cup \{ w \}$
		\EndIf
		\EndFor  
		\EndWhile
		\State \Return $\mathtt{NO\,SOLUTION} $
	\end{algorithmic} \caption{\label{alg:ucs} The ${\rm A}^*$ algorithm} 
\end{algorithm}

When the cost functional is simply the length of the path, as in equation (\ref{eq:length}), the canonical heuristic is the Euclidean distance between $\tilde{x}_0$ to $\tilde{x}_g$ which is the length of the optimal path in the absence of obstacles.
\begin{equation}\label{eq:length}
	c(\gamma)=\int_0^1 \Vert \gamma'(t) \Vert_2 \, {\rm d}t
\end{equation}
The Euclidean distance heuristic is specific to shortest path objectives, and may not be admissible for cost functionals other than (\ref{eq:length}). 
%
%

\section{The Landmark Heuristic}\label{sec:LH}

\begin{figure}
	\includegraphics[width=1.0\columnwidth]{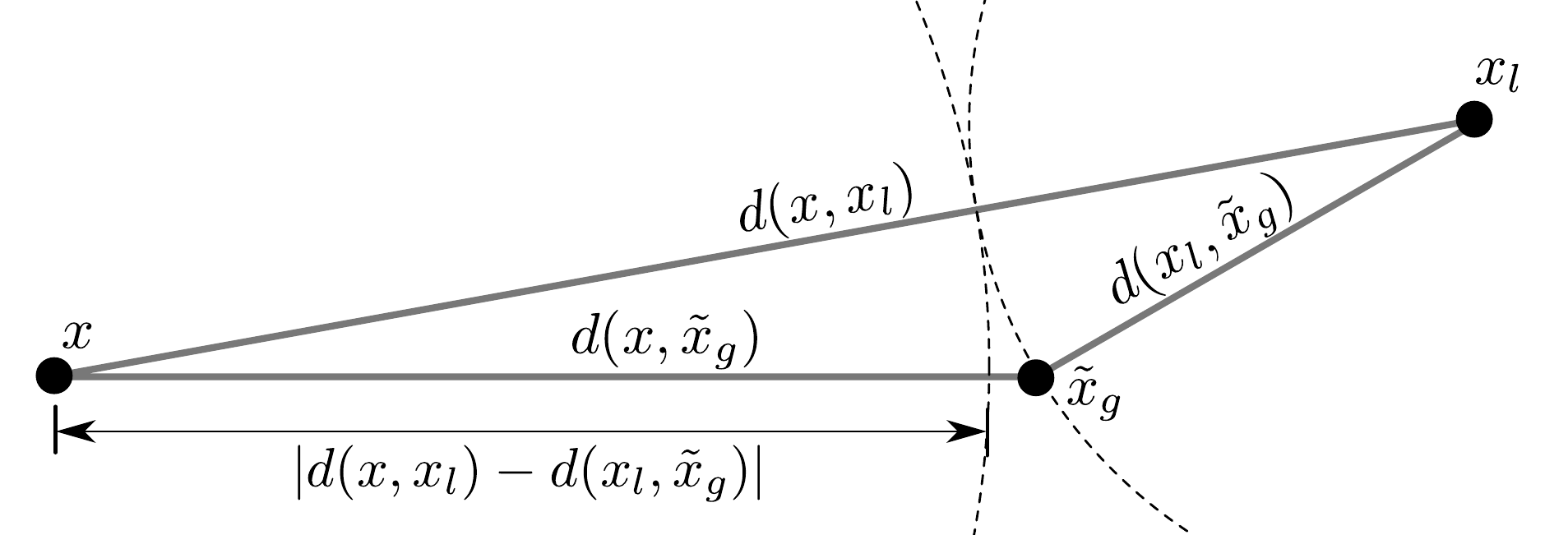}
	\caption{Geometric illustration showing how the triangle inequality can be rearranged to obtain a lower bound on the minimum cost path from $x$ to $\tilde{x}_g$.}\label{fig:triangle}
\end{figure}

The landmark heuristic is tailored to a particular graph and requires preprocessing the graph before it can be used in the \Astar algorithm.
The resulting heuristic is admissible regardless of cost functional and environment making it a very general approach to obtaining an admissible heuristic.
The idea behind the landmark heuristic is as follows:
Let $d:V_{\rm PRM}\times V_{\rm PRM} \rightarrow [0,\infty]$ be the function which returns the cost of a shortest path from one vertex of the graph to another; taking the value $\infty$ if no path exists. 
It follows from the definition that $d$ satisfies the triangle inequality.
Consider a vertex $x_l\in V_{\rm PRM}$ that will represent a \emph{landmark}. 
Rearranging the triangle inequality with $x_l$ and $\tilde{x}_g$ yields 
\begin{equation}\label{eq:triangle}
	|d(x,x_l)-d(x_l,\tilde{x}_g) |\leq d(x,\tilde{x}_g)\quad \forall x\in V_{\rm PRM}.
\end{equation} 
Thus, the left hand side of (\ref{eq:triangle}) is a lower bound on cost of the shortest path to $\tilde{x}_g$.
While computing $d$ explicitly would require solving the all-pairs shortest path problem, only the solution to the single-source shortest path problem from $\tilde{x_l}$ is required to evaluate (\ref{eq:triangle}). 
When the lower bound in (\ref{eq:triangle}) is evaluated at a vertex $x$ that lies on or near to the shortest path from $x_l$ to $\tilde{x}_g$ or vice-versa it provides a surprisingly close estimate of the minimum cost path from $x$ to $\tilde{x}_g$.
Figure \ref{fig:triangle} illustrates this lower bound.
However, obtaining an effective heuristic for all origin-destination pairs requires having a collection of landmarks $V_l\subset V_{\rm PRM}$. 
The landmark heuristic then leverages (\ref{eq:triangle}) for each landmark:
\begin{equation}\label{eq:landmark_heuristic}
h(x,x_g)=\underset{x_l\in V_l}{\max} \{|d(x,x_l)-d(x_l,\tilde{x}_g)|\}.
\end{equation}
%
To simplify the analysis presented in this paper, each landmark is an i.i.d. random variable selected from the uniform distribution on $V_{\rm PRM}$.
However, other selection rules can be used to improve the heuristic.

\subsection{Complexity of the Landmark Heuristic}
Generating the function $d(\cdot,x_l)$ for an individual landmark requires solving a single-source shortest path problem which can be accomplished with Dijkstra's algorithm in $O(|V_{\rm PRM}|\log(|V_{\rm PRM}|))$ time\footnote{This assumes the $\rm PRM$ graph is constructed using Algorithm \ref{alg:prm} which has $O(|V_{\rm PRM}|\log(|V_{\rm PRM}|)$ edges~\cite{karaman2011sampling}.} where $|\cdot|$ denotes the cardinality of a set.
Thus, the time complexity of constructing the heuristic is in $O(|V_{l}|\cdot|V_{\rm PRM}|\log(|V_{\rm PRM}|))$. 
From this observation it is clear that this heuristic is only useful in instances where the number of motion planning queries that will be evaluated on the ${\rm PRM}$ graph will be greater than $|V_l|$ since this many shortest path queries can be solved in the time required to construct the heuristic. 
Then evaluating the landmark heuristic (\ref{eq:landmark_heuristic}) requires looking up the optimal cost to a landmark  $2\cdot|V_l|$ times so the complexity of (\ref{eq:landmark_heuristic}) is linear in the number of landmarks.
Storing the cost of the shortest path to each vertex from a landmark for use in (\ref{eq:landmark_heuristic}) requires $O(|V_{\rm PRM}|)$ memory per landmark for a total memory requirement in $O(|V_l|\cdot|V_{\rm PRM}|)$. 
%
%
%
%

%

%

%

The next question is how many landmarks should be used? A natural choice is to select a fixed fraction of the \PRM vertices to be landmarks.
That is, $|V_l|=\kappa \cdot |V_{\rm PRM}|$ for some constant $\kappa$. 
This results in $O(|V_{\rm PRM}|^2)$ space required to store the heuristic's lookup tables in memory.
However, with just than 16 landmarks, the landmark heuristic has been observed to speed up routing queries by a factor of 9 to 16 on city to continent-scale road networks.
On a \PRM with a shortest path objective, this observation can be made precise as stated in the next result.
\begin{lem}\label{lem:subquadratic_thm}
	If the number of landmarks relative to the number of vertices is given by $|V_l|=\lambda\cdot|V_{\rm PRM}|$ for $\lambda\in(0,1]$, then
	\begin{equation}\label{eq:subquadratic_thm}
	\lim_{|V_{\rm PRM}|\rightarrow \infty} h(x,x_g)=d(x,x_g), 
	\end{equation}
	almost surely.
\end{lem}
The proof can be found in the appendix.
With increasing graph size and an arbitrarily small fraction of vertices assigned to landmarks, the landmark heuristic will converge to the solution of the all-pairs shortest path problem.

\subsection{Demonstration of the Landmark Heuristic}\label{sec:bug_trap}
\begin{figure}
	\includegraphics[width=1.0\columnwidth]{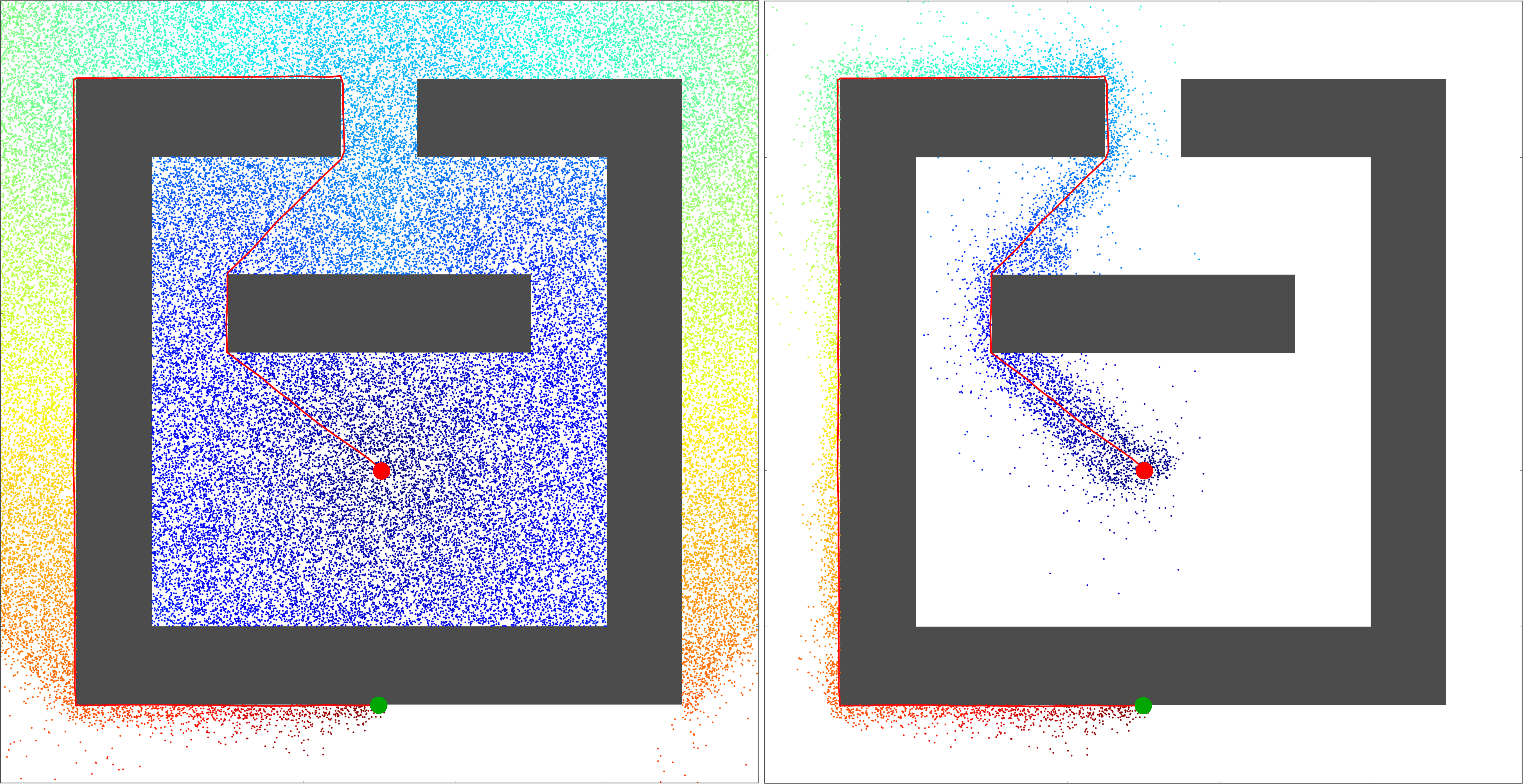}
	\caption{Shortest path from $\tilde{x}_0$ (red circle) to $\tilde{x}_g$ (green circle) on a ${\rm PRM}$ graph. Paths are computed with the Euclidean distance as a heuristic (left) and the landmark-based heuristic (right). Colored markers represent vertices examined in each search with color indicating the relative cost to reach that vertex.}\label{fig:but_trap}
\end{figure}

To demonstrate the advantages of using the landmark heuristic, it was compared with Dijkstra's algorithm and ${\rm A}^*$ with the Euclidean distance heuristic in a bug-trap environment. 
%
%
A \PRM was constructed in the bug trap environment according to Algorithm \ref{alg:prm} with a density of $1000$ vertices per unit area for a total of $69,272$ vertices. 
The Landmark heuristic was then constructed with $100$ landmarks ($0.14\%$ of vertices) obtained by randomly sampling from the vertices of the graph. 
Figure \ref{fig:but_trap} shows the environment and vertices expanded by the ${\rm A}^*$ algorithm using Euclidean distance as a heuristic and the landmark heuristic.
%
%
The \Astar algorithm with Euclidean distance heuristic required $58,145$ iterations and $351ms$ to find the shortest path; a marginal difference in performance in comparison to the $69,180$ iterations and $334ms$ required by Dijkstra's algorithm. 
In contrast, the \Astar algorithm with  the landmark heuristic required only $3,338$ iterations and $49ms$ to find the shortest path.
%
%
%

The results of this demo can be reproduced with the implementation of the landmark heuristic available in~\cite{landmark_implementation}.
%

%
\section{Evaluation in Randomized Environments}\label{sec:experiments}
%
%
Environments with randomly placed obstacles provides a simple and easily reproducible benchmark for motion planning algorithms~\cite{gammell2015batch,karaman2012high}.
In this paper, the degree of clutter in these randomly generated environments is quantified as the probability of the line segment connecting two randomly sampled points being contained in $X_{\rm free}$. 
%

%
\subsection{Environment Generation}\label{sec:environment}
A Poisson forest with intensity $\lambda$ of circular obstacles with radius $r$ is used as a random environment
This is simulated over a sample window $S=[-1,1]^2$ by sampling the number  of obstacles $N$ from the Poisson distribution
\begin{equation}
f_{N}(n)=\frac{(\lambda\mu(S))^{n}e^{-\lambda\mu(S)}}{n!},\label{eq:poisson}
\end{equation}
and then placing these obstacles randomly by sampling from uniform distribution on $S$. 
The subset of $S$ occupied by the circular obstacles is denoted $X_{\rm obs}$.
Then we select $X_{\rm free}=[-0.5,0.5]^2\setminus X_{\rm obs}$.
Embedding $X_{\rm free}$ in $S$ simplifies subsequent calculations by eliminating boundary effects of the sample window. 

\begin{figure}
	\begin{centering}
		\includegraphics[width=0.7\columnwidth]{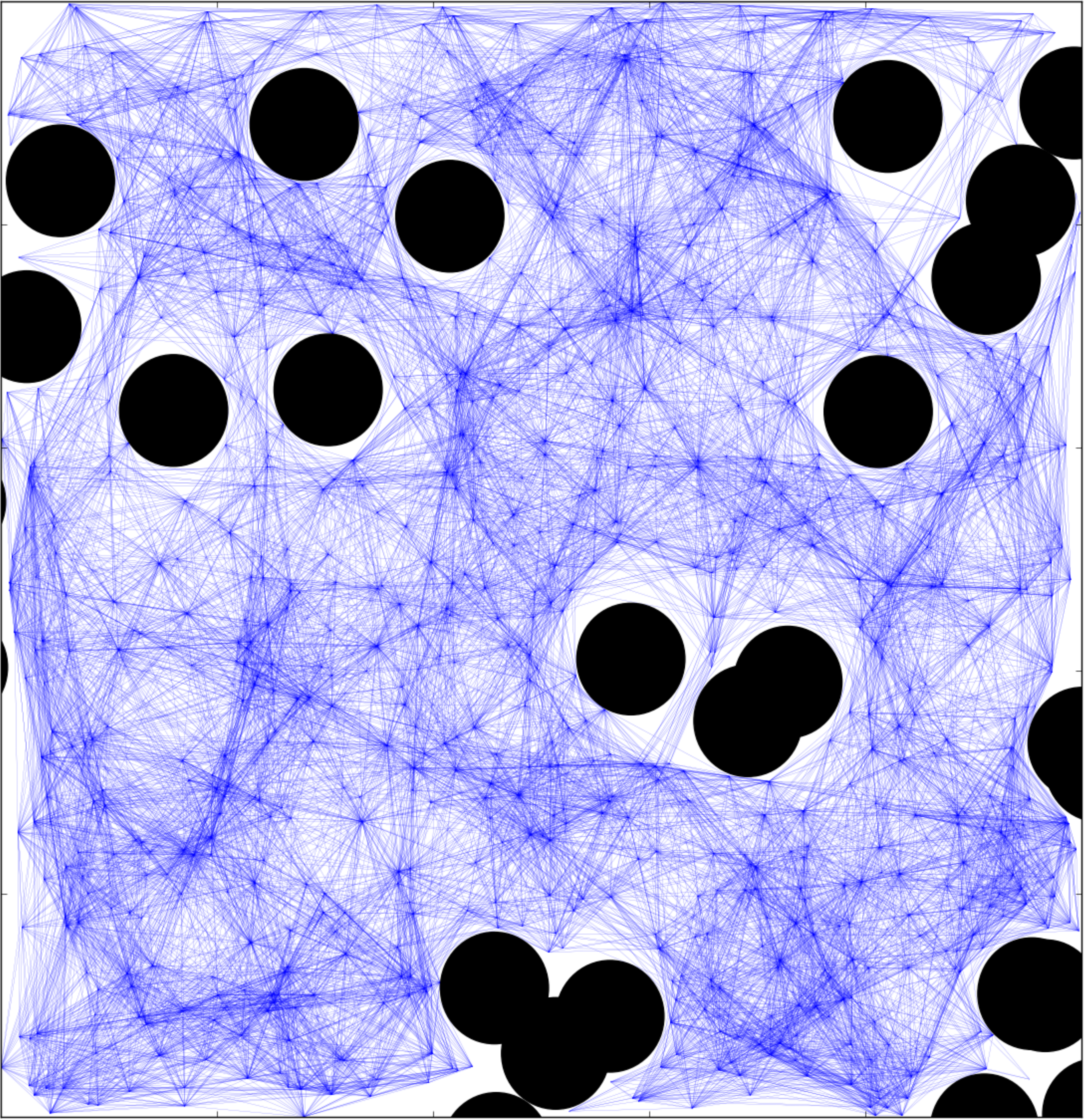}
		\par\end{centering}
	
	\caption{A ${\rm PRM}^*$ graph on a randomized environment with $\Pclear=0.5$. }
\end{figure}
Let $Z_1$ and $Z_2$ be independent random variables with the uniform distribution on $X_{\rm free}$,
and let $\texttt{clear}$ denote the event that the line segment connecting $Z_1$ and $Z_2$ remains in $X_{\rm free}$. 

The next derivation relates the obstacle intensity $\lambda $ to the marginal probability $\Pclear$.
Observe that a line segment intersects a circular obstacle of radius $r$ if and only if the circle of radius $r$ swept along this line segment contains the obstacle center.
If the obstacle is placed by sampling from the uniform distribution on $S$, the probability of collision is simply the ratio of the swept area of the circle along the line segment and the area of $S$.
Thus, conditioned on the number of obstacles $N$ and the points $Z_1,Z_2$, the probability of $\texttt{clear}$ is    
\begin{equation}
\mathbb{P}(\texttt{clear}|N,Z_{1},Z_{2})=\left(\frac{\mu(S)-\pi r^{2}-2r\Vert Z_{1}-Z_{2}\Vert}{\mu(S)}\right)^{N}.\label{eq:cond_prob}
\end{equation}
Then the marginal probability $\mathbb{P}(\texttt{clear})$ for a given obstacle intensity $\lambda$ can be calculated by combining (\ref{eq:poisson}) and (\ref{eq:cond_prob}) to obtain  
\begin{equation}
\begin{array}{l}
P(\texttt{clear})=\\
\underset{n\in\mathbb{N}}{\sum}\int_{X_{\rm free}}\int_{X_{\rm free}}\frac{P(\texttt{clear}|n,z_{1},z_{2})f_N(n)}{(\mu(\Xfree))^2}\,\mu(dz_{1})\mu(dz_{2}).
\end{array}
\label{eq:prob_A}
\end{equation}
In all of the numerical experiments of the next section random environments with obstacle radius $r=0.05$ were used.

\subsection{Numerical Experiments}

	Experiments were designed to evaluate how the effectiveness of the landmark heuristic varies with with the parameter $\mathbb{P}(\texttt{clear})$ and to validate Lemma \ref{lem:subquadratic_thm}. 
	To facilitate obtaining the results in a reasonable time, experiments were run in parallel on the central high-performance cluster EULER (Erweiterbarer, Umweltfreundlicher, Leistungsf\"ahiger ETH-Rechner) of ETH Z\"urich. Each compute node consists of two 12-Core  Intel Xeon E5-2680 processors with clock rates varying between 2.5-3.5 GHz.
	%
	%
	
	In the first set of trials a single random environment was sampled with $\Pclear=0.05$. 
	Three \PRM graphs were constructed on this environment with 40,000, 60,000 and 80,000 vertices.
	On each \PRM, 700 landmark heuristics were constructed, 100 each for  landmark quantities  $|V_l|\in\{10, 30,50,70,90,110,130\}$.
	Then for each landmark heuristic, a random shortest path query is solved using \Astar with the landmark heuristic. 	
	%
	
	%
	In the second set of trials, 20 logarithmically spaced values for the parameter $\Pclear$ from $0.01$ to $1.0$ were selected.
	For each of these values $100$ random environments were generated according to the construction outlined in Section \ref{sec:environment}.
	A \PRM with 100,000 vertices per unit area was constructed on each environment with  $100,000$ vertices per unit area.  
	Then for each \PRM, the $9$ landmark heuristics were constructed with landmark quantities $|V_l|\in\{10,30,50,70,90,110,130,150,170\}$.
	Finally, for  each of the $9$ landmark heuristics, $100$ shortest path queries were evaluated  on each \PRM using \Astar.  

\subsection{Results}
\begin{figure}
	\includegraphics[width=1.0\columnwidth]{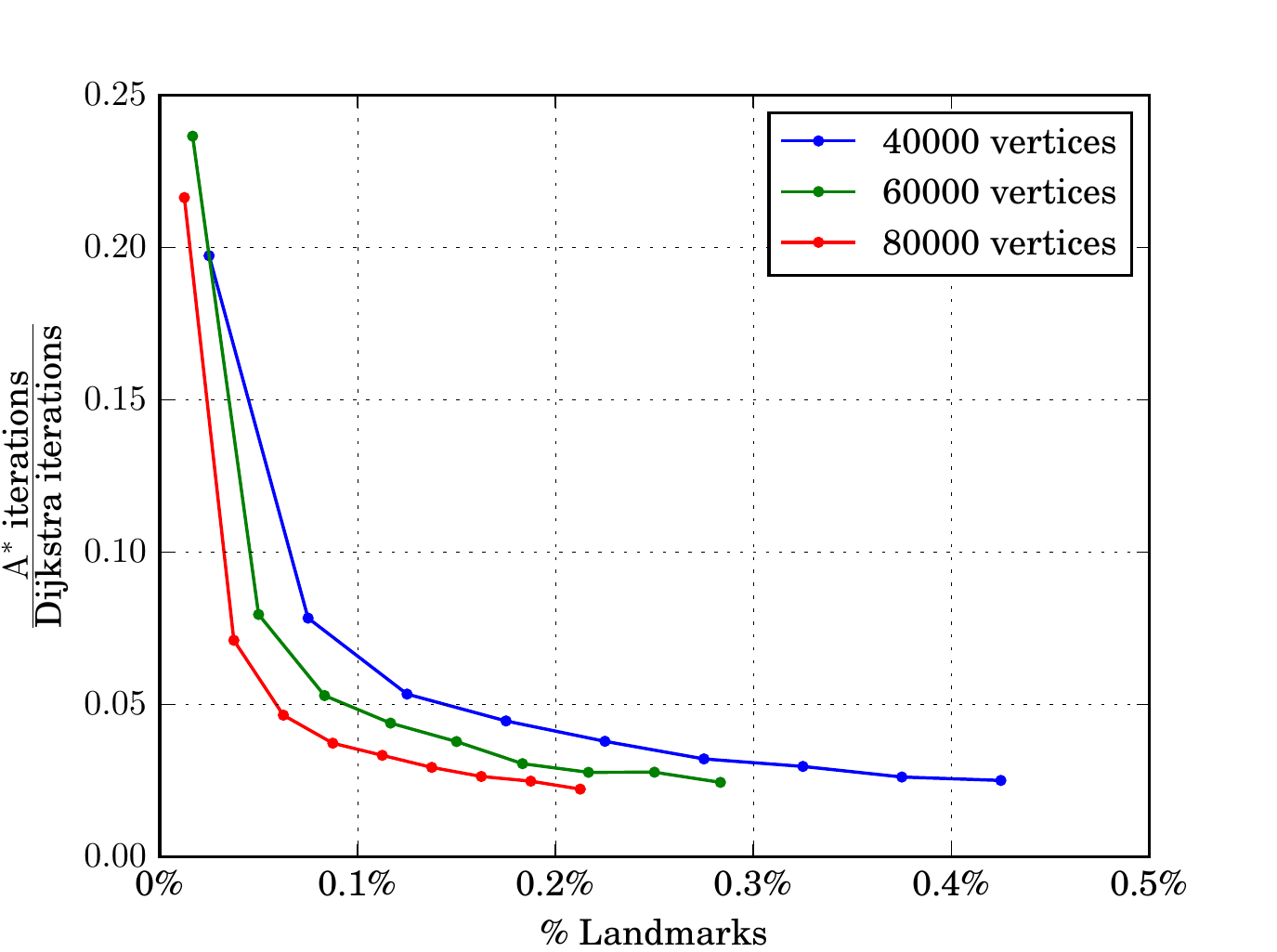}
	\caption{Effectiveness of the landmark heuristic increases with both the fraction of vertices assigned to landmarks \emph{and} the number of vertices in the graph.}\label{fig:subquadratic}
\end{figure}

\begin{figure}
	\includegraphics[width=1.0\columnwidth]{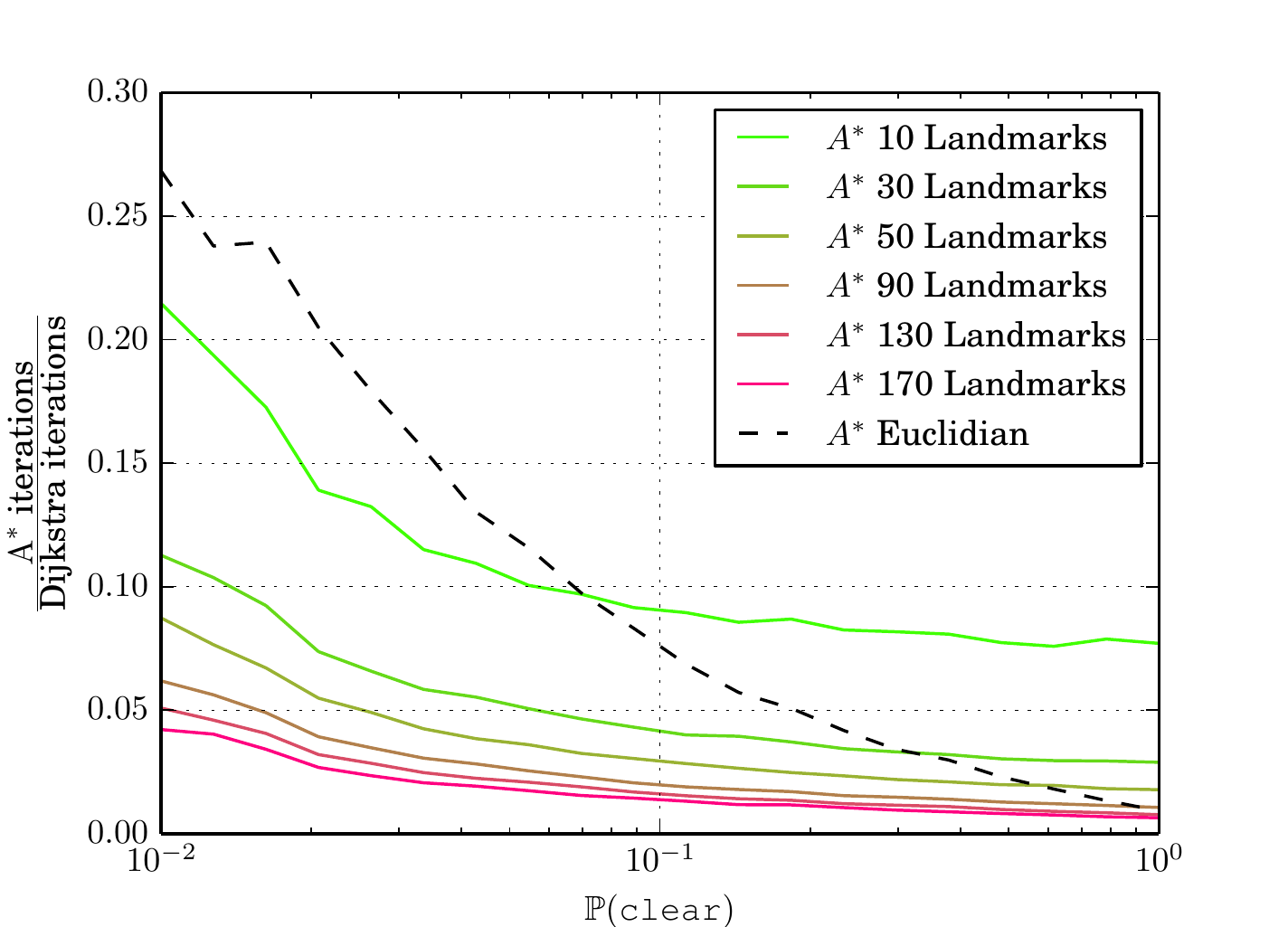}
	\caption{Effectiveness of the landmark heuristic in comparison to the Euclidean distance heuristic with varying degrees of clutter quantified by $\Pclear$.}\label{fig:iterations}
\end{figure}

The first experiment, summarized in Figure \ref{fig:subquadratic}, revealed how the effectiveness of the landmark heuristic varied with the fraction of vertices assigned to landmarks as well as with varying graph sizes.
We observed a rapid reduction in iterations required to find a solution relative to Dijkstra's algorithm with just $0.2\%$ of vertices assigned to landmarks.
Secondly, the number of iterations required to find a solution with \Astar relative to that of Dijkstra's algorithm decreased with increasing graph size. 
This validates Lemma \ref{lem:subquadratic_thm} since the number of iterations required by \Astar decreases with an improving estimate of the optimal cost to reach the goal.
In the second experiment we observed that the effectiveness of the Euclidean distance heuristic rapidly diminishes with increasing clutter, while the the landmark heuristic was much less sensitive to $\Pclear$. 
This is summarized in Figure \ref{fig:iterations} where the landmark heuristic reduced the number of iterations required to find a solution by a factor greater than 20 in highly cluttered environments whereas the Euclidean distance heuristic reduced the number of iterations by less than a factor of 3. 

\begin{figure}
	\includegraphics[width=1.0\columnwidth]{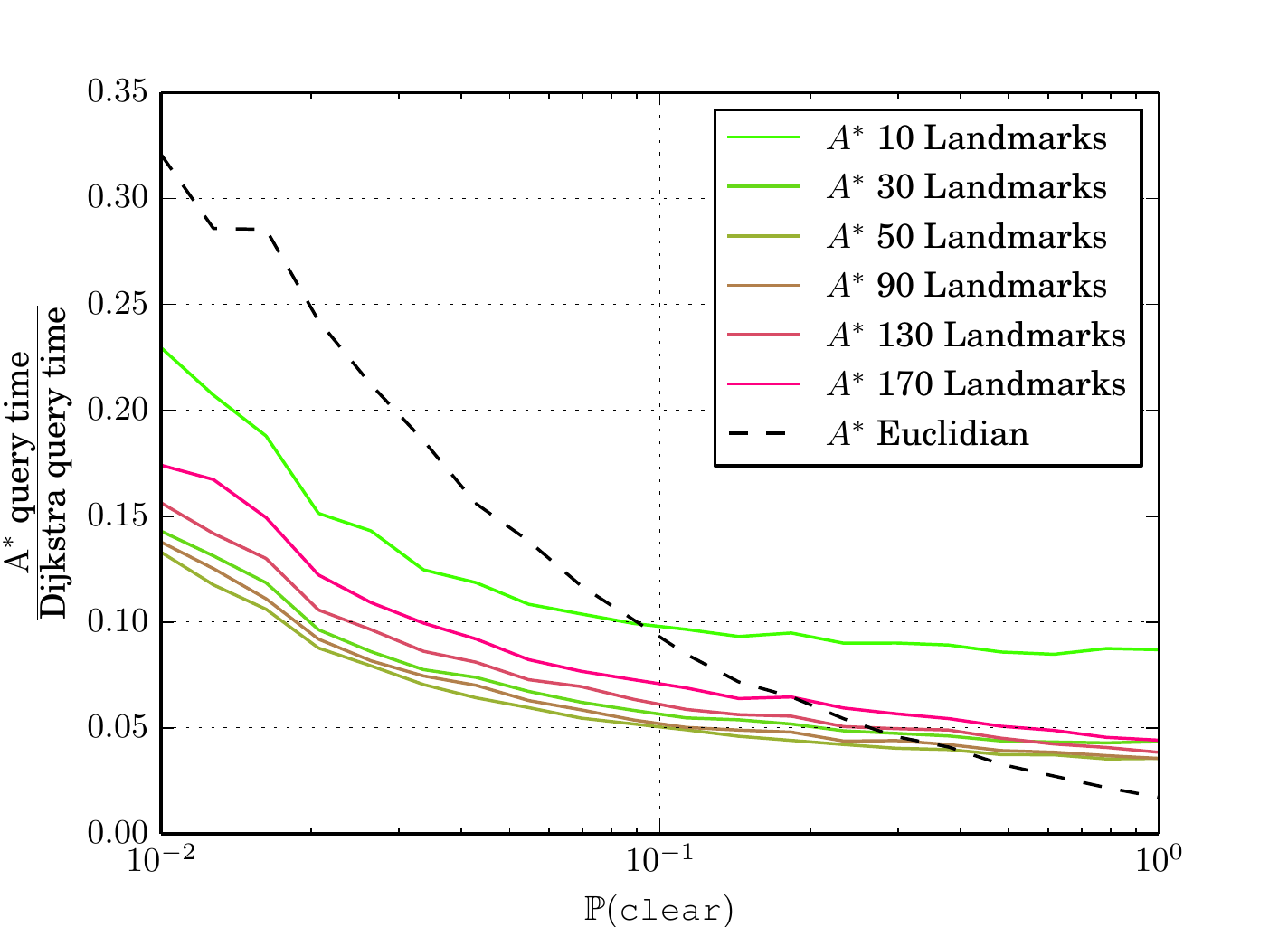}
	\caption{Running time of shortest path queries using the landmark heuristic and Euclidean distance heuristic normalized by the running time using Dijkstra's algorithm.}\label{fig:times}
\end{figure}

This experiment also showed the diminishing returns of increasing the number of landmarks in terms of iteration time.
Recall that evaluating the landmark heuristic in (\ref{eq:landmark_heuristic}) required checking the triangle equality for each landmark.
In Figure \ref{fig:times}  the average running time of the \Astar algorithm with the landmark heuristic reaches a minimum with with 50 landmarks. 
\begin{figure}
	\includegraphics[width=1.0\columnwidth]{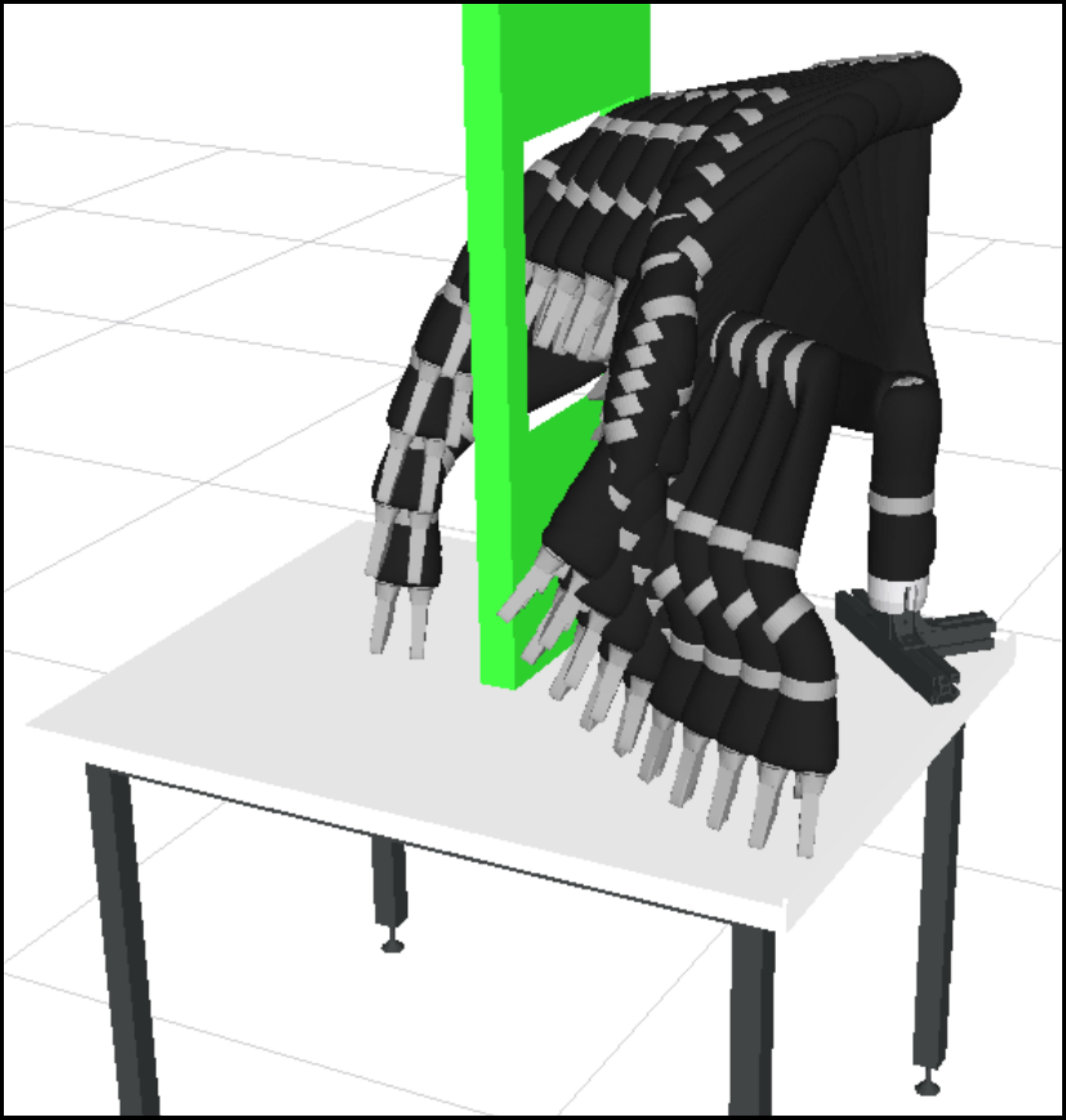}
	\caption{Still frames of the minimum mechanical work motion generated by the Jaco robotic arm when using a \PRM with 100'000 vertices and \Astar with the landmark heuristic.}\label{fig:jaco_trail}
\end{figure} 
\section{Robot Manipulator Example}\label{sec:jaco}
To demonstrate suitability of the landmark-based heuristic for realistic manipulator models, we use a model of the six degree of freedom Jaco manipulator by Kinova Robotics. 
To simulate a complex planning task the arm must find a collision free motion through a window terminating with the end effector near the ground to simulate reaching for an object.
The landmark heuristic was implemented in the Open Motion Planning Library (OMPL)~\cite{sucan2012the-open-motion-planning-library} and the problem was solved using the MoveIt~\cite{moveit} software tool.
Two planning objectives were considered for this problem, a shortest path objective and a minimum mechanical work objective. 
Motivation for using the shortest path objective is that the Euclidean distance is available as an admissible heuristic.
On the other hand, minimizing the mechanical work required to execute the motion is a more natural objective that is likely similar to the motion a human would use for the task.
The drawback to the latter objective is that there is no obvious heuristic to inform the \Astar search.
Since the landmark heuristic is admissible regardless of the objective it was applicable for this objective.

A $100,000$ vertex \PRM was constructed followed by the construction of a landmark heuristic with 50 landmarks.
A minimum work path was computed in $48$ iterations and $7.5ms$ using the landmark heuristic while Dijkstra's algorithm required $25,207$ iterations and $209.4ms$.
A shortest path was computed in $36$ iterations and $5.6ms$ using the landmark heuristic while using the Euclidean distance required $1,238$ iterations and $14.4ms$. 
Figure \ref{fig:jaco_trail} illustrates the minimum energy motion that was computed.

\section{Conclusion}\label{sec:conclusion}
The landmark heuristic is well known in the vehicle routing literature where it has been shown to reduce shortest path query times by a factor of 9 to 16 on city to continent-scale road networks.
Multi-query applications of the \PRM in robot motion planning have striking similarities with vehicle routing problems in road networks in that shortest path queries are evaluated repeatedly on a large graph.
The goal of this investigation was to evaluate the effectiveness of the landmark heuristic in robotic motion planning applications.
Since the heuristic is based on preprocessing the \PRM graph, our hypothesis was that its effectiveness would be independent of how densely cluttered the environment was---a useful feature for complex planning tasks.

To make this evaluation, we constructed a randomized environment parameterized by the probability that the line segment between two random points did not intersect obstacles.
The average case relative performance of the landmark heuristic relative to the Euclidean distance heuristic was then measured through numerous randomized trials.
Additionally, the performance of the landmark heuristic was evaluated on a manipulator arm model in a realistic planning scenario. 

The landmark heuristic was empirically observed to be less sensitive to environment clutter than the Euclidean distance heuristic.
For the range of parameters evaluated, the query times were reduced by a factor of 5 to 20 in comparison to Dijkstra's algorithm.
Secondly, a theoretical analysis showed that, with a fixed fraction of \PRM vertices assigned to be landmarks, the landmark heuristic converges to the optimal cost between any origin-destination pair with increasing graph size.
This analysis was then validated in our experimental results.
%

The landmark heuristic is an effective heuristic for querying large \PRM graphs.
In particular, it is more effective than the Euclidean distance heuristic in all but nearly obstacle free problem instances. 
However, the preprocessing time required to construct the heuristic makes it only suitable for multi-query applications where the heuristic will be used repeatedly on the same graph.
A valuable direction for future investigation would be an efficient update to the heuristic when small changes are made to the \PRM as a result of changes in the workspace.
\appendix{}
The proof of Lemma \ref{lem:subquadratic_thm} requires some additional notation. 
The symbol $\mu$ denotes the Lebesgue measure on $\mathbb{R}^d$ so that the uniform probability measure of a measurable subset $S$ of $\Xfree$ is given by $\mu(S)/\mu(\Xfree)$.
Since each landmark is an i.i.d. random variable with the uniform distribution on $\Xfree$, the set of landmarks $\{l_1,...,l_{|V_l|}\}$ can be viewed as a random variable on the product space, denoted $\Xfree^{|V_l|}$. 
The probability of $l_i\in S_i$ for subsets $S_i$ of $\Xfree$ is given by the product measure $m$:

\begin{equation}\arraycolsep=1.4pt\def\arraystretch{1.5}
\begin{array}{lll}
	\mathbb{P}\left(\{l_1,...l_{|V_l|}\}\in S_1\times ... \times S_{|V_l|}  \right) &=& m\left( S_1 \times ... \times S_{|V_l|} \right)\\
	&=&\prod_{i=1}^{|V_l|}\frac{\mu(S_i)}{\mu(\Xfree)}
	\end{array}
\end{equation}   
Next, an $\epsilon$-net on $\Xfree$ is a subset $\{z_1,...,z_k\}$ of $\Xfree$ such that 
\begin{enumerate}
	\item $\Xfree \subset \bigcup_{i=1}^M B_{\epsilon}(z_i) $,
	\item $B_{\epsilon /2}(z_i) \cap B_{\epsilon/2}(z_j)=\emptyset\quad \forall i\neq j$.
\end{enumerate}
Based on these two properties it is clear that the number of points $k$ making up an $\epsilon$-net on $\Xfree$ is bounded by 
\begin{equation}
\frac{\mu(\Xfree)}{\mu(B_{\epsilon}(\cdot))} \leq k \leq \frac{\mu(\Xfree)}{\mu(B_{\epsilon/2}(\cdot))}.
\end{equation}
Observe that not every $B_{\epsilon}(z_i)\cap \Xfree$ is convex since it may intersect the boundary of $\Xfree$. 
The the index set $\mathcal{K} \subset\{1,...,k\}$ will identify open balls of the $\epsilon$-net which have a convex intersection with $\Xfree$.
As the $\epsilon$-net becomes finer, a greater fraction of points will lie on the interior of $\Xfree$ with a distance to the boundary greater than $\epsilon$ so 
\begin{equation}\label{eq:cover}
	\lim_{\epsilon \rightarrow 0} \bigcup_{i\in \mathcal{K}} B_{\epsilon}(z_{i} )=\Xfree.
\end{equation} 

\begin{proof}[Proof (Lemma \ref{lem:subquadratic_thm})]
	Consider the $\epsilon$-net described above with $\epsilon=r/2$, half the connection radius of the \PRM.
	Note that every vertex in $B_{r/2}(z_i)$ is connected by a line segment for $i\in\mathcal{K}$.
	The probability that the landmarks $\{l_1,...,l_{|V_l|}\}\cap B_{r/2}(z_i)=\emptyset$ for some $i\in\mathcal{K}$ can be written as
	\begin{equation}\label{eq:mess}
	\begin{array}{l}
		\mathbb{P}\left( (l_1,...,l_{|V_l|}) \in \bigcup_{i\in \mathcal{K}} (B_{r/2}^c(z_i))^{|V_l|} \right) \\ \leq \sum_{i\in \mathcal{K}} m((B_{r/2}^c(z_i))^{|V_l|})\\ \leq
		\frac{\mu(\Xfree)}{\mu(B_{r/4})} \cdot\left(1-\frac{\mu(B_{r/2}(z_i))}{\mu(\Xfree)}\right)^{|V_l|} 
	\end{array}
	\end{equation}
	By inserting the expression for $r$ in Algorithm \ref{alg:prm} and replacing $|V_l|$ with $\lambda\cdot |V_{\rm PRM}|$, the last expression in (\ref{eq:mess}) simplifies to
	\begin{equation}\label{eq:p_conv_zero}
	\alpha \cdot\frac{|V_{\rm PRM}|}{\log(|V_{\rm PRM}|)} \cdot \left(1-\beta\cdot\frac{\log(|V_{\rm PRM}|)}{|V_{\rm PRM}|} \right)^{\lambda \cdot |V_{\rm PRM}|},
	\end{equation}	
	for constants $\alpha,\beta\in(0,1)$ that depend only on the dimension $d$ of $\Xfree$. 
	One can readily verify that this expression converges to zero as $|V_{\rm PRM}|\rightarrow \infty$. 
	Thus, the probability that there is at least one landmark in each $B_{r/2}(z_i)$ for $i\in \mathcal{K}$ converges 1.  
	It follows that every vertex $x\in V_{\rm PRM}\cap\left(\bigcup_{i\in\mathcal{K}} B_{r/2}(z_i)\right)$ has a landmark as a neighbor almost surely as $|V_{\rm PRM}|\rightarrow \infty$.
	Therefore, for at least one landmark $l^*$, the optimal cost from $x$ to $l^*$ satisfies $d(x,l^*)\leq r$. 
	Thus,
	\begin{equation}\label{eq:convergence_bound}
	\begin{array}{rcl}
	
	h(x,x_g)&=&\underset{x_l\in V_l}{\max} \{|d(x,x_l)-d(x_l,\tilde{x}_g)|\}\\
	&\geq& d(x,l^*)-d(x_g,l^*)
	\end{array}
	\end{equation}
	Expanding $d(x,l^*)$ with the triangle inequality between $x$, $l^*$, and $x_g$ yields
	\begin{equation}
	\begin{array}{rcl}
	h(x,x_g)&\geq& d(x,x_g)-d(x_g,x_n)-d(x_g,x_n)\\
	&\geq& d(x,x_g)-2r
	\end{array}
	\end{equation}  
	Combining (\ref{eq:triangle}) and (\ref{eq:convergence_bound}) we have $d(x,x_g)-2r\leq h(x,x_g) \leq d(x,x_g)$, and since $r\rightarrow 0$ as $|V_{\rm PRM}|\rightarrow \infty$ we obtain 
	\begin{equation}
	\lim_{|V_{\rm PRM}|\rightarrow \infty} h(x,x_g)=d(x,x_g), 
	\end{equation}
	on $\bigcup_{i\in\mathcal{K}} B_{r/2}(z_i)$.
	The desired result (\ref{eq:subquadratic_thm}) then follows in light of (\ref{eq:cover}).
\end{proof}

\bibliographystyle{ieeetr}
\bibliography{references}

\begin{thebibliography}{10}

\bibitem{kavraki1996probabilistic}
L.~E. Kavraki, P.~Svestka, J.-C. Latombe, and M.~H. Overmars, ``Probabilistic
  roadmaps for path planning in high-dimensional configuration spaces,'' {\em
  IEEE transactions on Robotics and Automation}, vol.~12, no.~4, pp.~566--580,
  1996.

\bibitem{lavalle2001randomized}
S.~M. LaValle and J.~J. Kuffner, ``Randomized kinodynamic planning,'' {\em The
  International Journal of Robotics Research}, vol.~20, no.~5, pp.~378--400,
  2001.

\bibitem{hsu1997path}
D.~Hsu, J.-C. Latombe, and R.~Motwani, ``Path planning in expansive
  configuration spaces,'' in {\em Robotics and Automation, 1997. Proceedings.,
  1997 IEEE International Conference on}, vol.~3, pp.~2719--2726, IEEE, 1997.

\bibitem{murray2016robot}
S.~Murray, W.~Floyd-Jones, Y.~Qi, D.~Sorin, and G.~Konidaris, ``Robot motion
  planning on a chip,'' in {\em Robotics: Science and Systems}, 2016.

\bibitem{marble2013asymptotically}
J.~D. Marble and K.~E. Bekris, ``Asymptotically near-optimal planning with
  probabilistic roadmap spanners,'' {\em IEEE Transactions on Robotics},
  vol.~29, no.~2, pp.~432--444, 2013.

\bibitem{goldberg2005computing}
A.~V. Goldberg and C.~Harrelson, ``Computing the shortest path: A search meets
  graph theory,'' in {\em Proceedings of the sixteenth annual ACM-SIAM
  symposium on Discrete algorithms}, pp.~156--165, Society for Industrial and
  Applied Mathematics, 2005.

\bibitem{karaman2011sampling}
S.~Karaman and E.~Frazzoli, ``Sampling-based algorithms for optimal motion
  planning,'' {\em The International Journal of Robotics Research}, vol.~30,
  no.~7, pp.~846--894, 2011.

\bibitem{landmark_implementation}
B.~Paden, Y.~Nager, and E.~Frazzoli, ``Landmark guided probabilistic roadmap
  queries,'' 2017.
\newblock Available at:
  \url{https://github.com/bapaden/Landmark_Guided_PRM/releases/tag/v0}.

\bibitem{gammell2015batch}
J.~D. Gammell, S.~S. Srinivasa, and T.~D. Barfoot, ``Batch informed trees
  (bit*): Sampling-based optimal planning via the heuristically guided search
  of implicit random geometric graphs,'' in {\em 2015 IEEE International
  Conference on Robotics and Automation (ICRA)}, pp.~3067--3074, IEEE, 2015.

\bibitem{karaman2012high}
S.~Karaman and E.~Frazzoli, ``High-speed flight in an ergodic forest,'' in {\em
  Robotics and Automation (ICRA), 2012 IEEE International Conference on},
  pp.~2899--2906, IEEE, 2012.

\bibitem{sucan2012the-open-motion-planning-library}
I.~A. {\c{S}}ucan, M.~Moll, and L.~E. Kavraki, ``The {O}pen {M}otion {P}lanning
  {L}ibrary,'' {\em {IEEE} Robotics \& Automation Magazine}, vol.~19,
  pp.~72--82, December 2012.
\newblock \url{http://ompl.kavrakilab.org}.

\bibitem{moveit}
I.~A. Sucan and S.~Chitta, ``Moveit!,'' 2016.

\end{thebibliography}

\end{document}